\documentclass{article}

\usepackage[final]{nips_2018}

\usepackage[utf8]{inputenc} % allow utf-8 input
\usepackage[T1]{fontenc}    % use 8-bit T1 fonts
\usepackage{hyperref}       % hyperlinks
\usepackage{url}            % simple URL typesetting
\usepackage{booktabs}       % professional-quality tables
\usepackage{amsfonts}       % blackboard math symbols
\usepackage{nicefrac}       % compact symbols for 1/2, etc.
\usepackage{graphicx}  %Required

\title{Flexible and Scalable State Tracking Framework for Goal-Oriented
Dialogue Systems}

\author{Rahul Goel, Shachi Paul, Tagyoung Chung \\
   \textbf{Jeremie Lecomte, Arindam Mandal,  Dilek Hakkani-Tur}  \\
  Amazon Alexa AI \\
  \texttt{ \{goerahul, shachp,  tagyoung, lecomtj, arindamm, hakkanit\}@amazon.com} }

\begin{document}

\maketitle

\begin{abstract}

Goal-oriented dialogue systems typically rely on components
specifically developed for a single task or domain. This limits such
systems in two different ways: If there is an update in the task
domain, the dialogue system usually needs to be updated or completely
re-trained. It is also harder to extend such dialogue systems to
different and multiple domains. The dialogue state tracker in conventional
dialogue systems is one such component --- it is usually designed to
fit a well-defined application domain. For example, it is common for a
state variable to be a categorical distribution over a
manually-predefined set of entities~\citep{henderson2013deep},
resulting in an inflexible and hard-to-extend dialogue system. In this
paper, we propose a new approach for dialogue state tracking that can
generalize well over multiple domains without incorporating any
domain-specific knowledge. Under this framework, discrete dialogue
state variables are learned independently and the 
information of a predefined set of possible values for dialogue state
variables is not required.  Furthermore, it enables adding
arbitrary dialogue context as features and allows for
multiple values to be associated with a single state variable.  These
characteristics make it much easier to expand the dialogue state space.
%or apply the same framework to completely different domains.
We evaluate our framework using the widely used dialogue state tracking challenge data
set (DSTC2) and show that our framework yields competitive results with other
state-of-the-art results despite incorporating little domain knowledge. We also
show that this framework can benefit from widely available external resources
such as pre-trained word embeddings. 

\end{abstract}

\section{Introduction} \label{sec:introduction}

With the rise of digital assistants such as Alexa, Cortana, Siri, and Google
Assistant, conversational interfaces are increasingly becoming a part of
everyday life. Such conversational interfaces can be broadly divided into
chit-chat systems and task-oriented systems~\cite{}. Chit-chat systems are
designed to be engaging to users so as to carry on natural and coherent
conversations. Task-oriented systems, on the other hand, are designed to help
users accomplish certain goals such as booking an airline ticket or making a
reservation. With the advent of pervasive use of deep learning, there has been an
increase in end-to-end learning of chit-chat
systems~\citep{serban2016building,yao2016attentional,shao2017generating}
employing varieties of sequence-to-sequence networks~\citep{sutskever2014sequence}.
Compared to task-oriented dialogue, end-to-end learning has been more widely
adopted for chit-chat dialogue systems due to the following reasons:
\begin{itemize}
\item 
\textbf{Availability of data}: Training dialogue systems end-to-end requires
large data sets, i.e, millions of dialogues of human-human conversations are
available from social media sites such as Reddit and Twitter that are useful for
training chit-chat systems.  In comparison, there are far fewer task-oriented
dialogue data sets available and their sizes are much
smaller. SNIPS~\footnote{https://github.com/snipsco} and
ATIS~\citep{hemphill1990atis} are examples of task-oriented dialogue data sets
and they incldue less than a thousand dialogues.
\item
\textbf{Lack of hard constraints}: Due to the open-ended nature of chit-chat
dialogue, the system only needs to be engaging and coherent. In task-oriented
dialogue systems, there are additional requirements such as being able to ask
for correct information in non-repeatative manner and being able to perform
correct actions that are in line with the user's intents.
\item
\textbf{No connection to downstream applications}: Chit-chat dialogue systems do
not necessarily have to interface with downstream applications or databases,
although recently, there has been a focus on knowledge grounded interactions
(e.g., DSTC 6 and 7~\citep{hori2018overview}). On the other hand, goal-oriented
dialogue systems must be able to interface with external systems with discrete
input representations, which adds additional difficulties to end-to-end
learning.
\end{itemize}

Due to these reasons, though there are only a few examples of complete end-to-end
learning of goal-oriented dialogue systems~\citep{dhingra2016towards,
liu2017end, liu2018dialogue, zhao2016towards}, such approaches remain uncommon,
especially in commercially deployed systems. Most traditional goal-oriented
dialogue systems are built as a pipeline with modules for spoken language
understanding, state tracking, and action selection. Typically, each of these
modules is specifically designed for intended domains with manually chosen
label spaces. Thus, scaling these components to many different domains or more
complex use cases can be very challenging.

In a multi-turn dialogue, the system has to be able to reason over the dialogue
states to ask necessary questions, inform the users with valid information, or
perform sequences of actions needed to achieve user's goals. To accommodate these
requirements, dialogue states are generally defined as discrete variables that
are symbolic so as to allow the system to reason over them. 
% NOTE: You could remove the following sentence, but I think this is an important reason:
Furthermore, backend APIs used to accompish user's goals require such discrete input, for example, SQL or SPARQL queries.

Despite these challenges, recently there has been a push for representing
dialogue states with neural networks similar to chit-chat dialogue systems. One
key advantage being the latent dialogue state representation~\citep{liu2017end}.
With end-to-end learning, the system can learn a distributional state
representation as well as reason over it. However, there are few shortcomings to
this approach:
\begin{itemize}
\item
Lack of annotated data: Automatically inferring dialogue states may require
thousands of annotated conversations to learn simple actions such as updating a
value associated with an entity.
\item
Lack of an easy way to inject constraints: In practical settings, placing hard
constraints on how dialogue system operates might be necessary, e.g., asking for
pin-code before making a voice purchase.
\item
Lack of interpretability: Due to the absence of symbolic dialogue states, it may be
hard to explain why the system chose to perform a particular action. This kind
of interpretability might be essential for commercial applications.
\end{itemize}

In this paper, we propose a flexible and extensible dialogue state tracking
framework that still relies on updating values of discrete variables but the
framework can be applied to new domains and new use cases much more easily.
Our system has the following desirable properties: First, there is no explicit
rules encoded for state variable updates. For example, values for state
variables are not confined to pre-set possible values. 
% soften the language here. "is" -> "can be"
Second, each state variable can be learned independently from each other. This
makes it easier to add new state variables to the system thus making the extension of the
framework to new use cases easier. Third, our framework allows for multiple
values to be associated with a state variable resulting in richer
representations. 
% shall we give an example here? for example, the user may say at 4 or 5pm
 In most other frameworks, this is not possible. This makes our
framework more flexible such that representing more complex state becomes
easier. All these lead to increased scalability in terms of increasing domain
complexity or adapting the framework to multiple different domains.

In our experiments with the DSTC2 corpus~\citep{henderson2014second}, we show that this framework yields
competitive results to state-of-the-art on this data~\citep{williams2016dialog}.
We show that even without explicit output space definition, our results are
competitive. We also show that models benefit from pre-trained word embeddings,
which opens doors for incorporating dynamic word embeddings to state tracking
models. Dynamic word embeddings have been shown to improve multiple natural language processing
applications~\citep{devlin2018bert}. We believe this work is a step towards more
scalable end-to-end learning of task-oriented dialogue systems, which we discuss
in the following sections.

\section{Methodology} \label{sec:methodology}

\begin{figure}
%[b]
  \centering
  \includegraphics[width=1.1\textwidth]{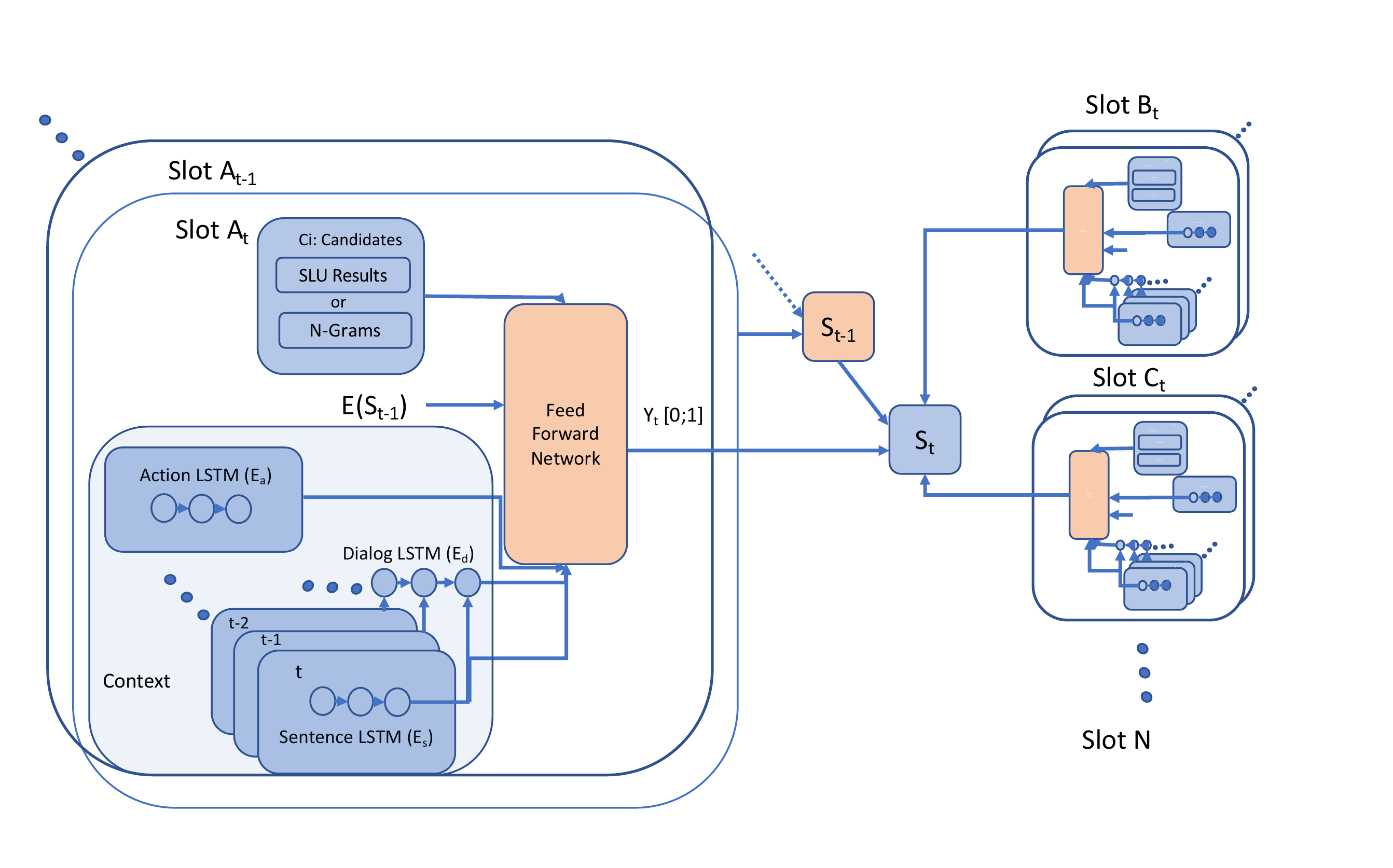}
  \caption{Dialogue state tracking framework. The main inputs to the system are
candidate values($C_i$) for dialogue state variables, previous state ($E(S_{t-1})$),
and various context($E_a$ and $E_d$) including the current user utterance($E_s$). The output is a binary
number indicating whether a candidate value under consideration is a value for
the current state variable.}
  \label{fig:framework}
\end{figure}

The dialogue state tracking task involves inferring the current state of
dialogue based on the conversation that has happened so far. A state could be
represented as a set of variables~(slot-value pairs or API name with its arguments)
associated with users' goals and the system capabilities.
%% System actions are generally represented as a set of dialogue acts
%% and slot value combinations or APIs and its respective argument values.
In this
setting, the goal of the system becomes predicting either the right dialogue act
and slot value combinations that form the system's response or calling the
correct APIs with correct arguments to help the user achieve his or her goal.
%% Thus, our state tracking model is a system designed to infer the slot-value
%% pairs that make up a user goal given the current state of the dialogue.

More concretely, the dialogue state consists of user goals and slots associated
with the goal. The goals and slots depend on the domain ontology and are
typically handcrafted. The DSTC2 dataset belongs to the restaurant domain. It
consists of only one goal which is to find a restaurant and there
are three slot types: \textit{cuisine}, \textit{area}, and \textit{pricerange}.
In this case, the dialogue state is represented by a triple with corresponding
values for each slot type. When a user says ``I want a cheap indian
restaurant,'' the dialogue state can be represented as a value triple:
\textit(\verb+indian+, \verb+none+, \verb+cheap+) corresponding to the slots
\textit{cuisine}, \textit{area}, and \textit{pricerange}.

Given a user goal, we adopt an open-vocabulary scoring model for each
slot type that needs to be tracked for the goal. Similar to
\citep{rastogi2018multi}, the input to our model is a set of
candidates $C_i \in 1..C$ that could be a value for each slot type, a
slot type $T_i \in 1..N$, and conversation context $D$.  Using
$n$-grams present in an utterance as the candidate values makes
either spoken language understanding (SLU) system or the named entity
recognition (NER) system unnecessary and we can do a joint SLU and
state tracking. However, if there is SLU or NER for the domain
available, we can use it to reduce the search space. Our model
produces a binary classification decisions for each combination of a
candidate value and a slot type. A positive class corresponds to the
slot value being the correct value for the slot type. Note that we do
not place any constraints on how many values can be associated with
the slot type, hence we can represent slots with multiple values. For
example, food=\{chinese, thai\} in the user utterance: ``I am looking
for chinese or thai restaurants''.

Figure~\ref{fig:framework} shows the overall framework of this approach. Note the
following characteristic of this model. First, each decision, i.e., whether a
candidate value can be used to update the current state, is made completely
independently from all other decisions. Second, each decision is made
independently from other slot types. Third, the model only makes a binary decisions
per candidate value and it does not rely on a pre-defined output space. Fourth,
the model can be configured to consider varieties of dialogue context.

With the available input, we build a fully-connected neural network classifier. For a
given candidate and a slot type, we minimize the cross entropy loss as shown
below:
\begin{equation}
  \min \sum_{k=1}^{N} -\sum_{i=1}^{C}\log\left(P\left(\hat{y}|C_i,T_k,D\right)\right).
\end{equation}
% NOTE: we need to define the variables in the equation here:
where $C_i$ denotes the candidate set of values, $T_k$ denotes the set of slot types and $D$ denotes the conversation context (Equation~\ref{context}).
% let's make S_k something else, as we use S_t to denote the latent space.

Given a user turn $i$ , we construct a candidate set for that turn. A candidate
set is an open set consisting of possible slot values for each slot type. In a
typical dialogue system this could be constructed from the output of SLU system
augmented with additional values obtained using simple rules~(such as business
logic). For our experiments, we create two kinds of candidate sets for each
turn:
\begin{enumerate}
\item
$N$-gram candidate set: This consists of the $n$-grams obtained from the user
utterance. We experiment with unigrams ($O(n)$) or bigrams ($O(2*n)$) in this
paper.
% NOTE: bigrams is O(2*n), but all n-grams are O(n^2). We need to correct the above.
\item
SLU candidate set: The DSTC2 data provides us with SLU results for each of the
user utterances, which are noisy. We experiment with just using the SLU results
for each slot type as our candidate set.
\end{enumerate}
Furthermore, we add a value \verb+dontcare+ to each of these candidate sets to
allow for no preference for each of the slot type. We construct our training
data by assigning the candidates a value of $1$ or $0$ for each slot type based
on the ground truth dialogue state. The system starts with a default state for a
given action or an API. After each user utterance, we update our dialogue state
with candidates that are predicted as positive. Based on the system design, various
update strategies or constraints can be incorporated in the dialogue state
update step. For example, if we want to enforce the constraint that one slot can
have only one value, we can select the candidate with the highest score from the
pool of positive candidates.

The dialogue history context features are flexible and we can easily add new context
features by appending them to the existing context vector. For our experiments we
use the following context features at each user turn $i$. 
\begin{enumerate}
\item Sentence Encoding ($E_s$) using LSTM for the current sentence
  $LSTM^{sentence}(Sent_i)$
\item Hierarchical LSTM ($E_d$) over past sentences to encode the dialogue context
  $LSTM^{dialogue}[LSTM^{sentence}_{1..i-1}(Sent_i)]$
\item LSTM ($E_a$) over system dialogue acts $LSTM^{dialogueAct}_{1..K}(DA_k)$
\item Previous state encoding ($E(S_{t-1})$):  For each slot type we
  learn a dense $S \times E_s$ matrix where $S$ is the number of states and
  $E_s$ is the state embedding dimension.
  We do an embedding lookup over the previous
  state and use it as a feature.% $Emb(State_{t-1})$.
\end{enumerate}
We concatenate all of these features into a context feature vector $D$. For every
candidate and slot type we have:
\begin{equation} \label{context}
  D = [E_s;E_d;E_a;E(S_{t-1})]
\end {equation}
\begin{equation}
  \hat{y} = F(C_i, D).
\end{equation}

%% By decomposing the problem in the aforementioned way, our system inherently
%% becomes scalable and easy to adapt for new slot types.
%% Since a binary decision is made for each candidate-slot pair, the output space
%% is not limited by the slot values in the training set and also allows for
%% multiple values for a given slot.
%% It also makes expanding the state
%% variable space easier.
We also show through experiments that our performance
% Rahul: Should we move some of this to conclusion results?
improves when we use pretrained embeddings for the $n$-gram candidate set, such a
system could take advantage of contextual language models such as
ELMO~\citep{peters2018deep}, which has been shown to improve many different
natural language processing tasks. Our system is also computationally efficient
due to the binary classification decision per slot compared to a softmax over
slot-values. Due to our design we inherently support multi-word expressions and
multiple values per slot which is hard to support if the output space is
the set of predefined slot values.

\section{Experiments} \label{sec:experiments}

\subsection{Data} \label{sec:data} 

We evaluate our model on the DSTC2~\citep{henderson2014second} dataset.
Dialogues in DSTC2 are in the restaurant search domain wherein a user can search
for a restaurant in multiple ways by selectively applying constraints on a set
of ``user-informable'' slots. The values taken by these slots are either
provided by the user in an utterance or are a special value --- \verb+dontcare+,
which means the user has expressed that he/she has no preference. The slot value
can also be \verb+none+ which indicates the user has not provided any
information for this slot type yet.  When representing system actions, we use
the dialogue acts and slot types but do not use the slot values (except
\verb+dontcare+). Hence the system action \verb+inform(price = cheap)+, is
represented as \verb+inform(price)+.  Additionally, the dataset also contains
the results obtained from the SLU system, which we use as our candidate set in
some experiment configuration.

Similar to \citet{liu2018dialogue}, we combine the provided training and
development datasets and then perform a 90/10 split for validation during
training. The test dataset is used as it is provided in the data set. Statistics
of this dataset are summarized in the Table~\ref{data1}.

\begin{table}[h]
 \caption{Statistics of the DSTC2 dataset.}
 \label{data1}
  \centering
  \begin{tabular}{lrrrr}
    \toprule
    Size of train and dev / test dialogues & $2118$ / $1117$ \\
    Average number of turns per dialogue & $7$ \\
    Number of values for \textit{cuisine} / \textit{area} / \textit{price} & $74$ /
$5$ / $3$ \\
    \bottomrule
   \end{tabular}
\end{table}

\subsection{Configuration}

We use word embeddings size of $300$. Hidden dimension for sentence LSTM is $128$
and $256$ for the dialogue LSTM. For the dialogue acts LSTM, we use embeddings
of size $50$ and LSTM hidden size of $64$. Our state embeddings ($E_s$) is of size
$16$. We use ADAM for
optimization~\citep{kingma2014adam} with learning rate of $0.001$ for all our
experiments. We use mini-batch of size $128$ during training. We limited our
context to previous five utterances to make batch training more efficient. We use
dropout of $0.5$ between fully connected layers. We also explore using FastText
pretrained word embeddings~\citep{bojanowski2016enriching} instead of random
initialization for our word embeddings. Due to the set up of our framework, our
training set is very imbalanced (too few positive classes). We up-weighted the
positive class by a factor of $8$ in the cross entropy loss to mitigate this
issue. We use $2$ fully connected layer of size $256$ and $16$ in succession for
final classification. We also do an ensemble by majority voting on the joint state
for 4 runs. 

\subsection{Results} \label{sec:results}

For the DSTC2 data set, we used a simple dialogue state update strategy, for
every candidate and slot type if the candidate is classified as positive, we
update the dialogue state using that candidate's value in the order it appeared
in the sentence. The goal and joint goal accuracy results on the two candidate
set settings~(SLU and N-gram candidates) are shown in Table~\ref{dialogue1}. We
report the standard evaluation metrics for this dataset.

\begin{table}[h]
  \caption{Dialogue state tracking accuracy performance on the DSTC2 test
    set. We compare candidates obtained using unigrams and bigrams versus
    candidates obtained from the SLU system. In the first case, the model is
    doing joint state tracking and SLU. We report the per goal and joint
    accuracy. Our result is the average of four runs of training and the
    standard deviation is in the brackets.}
  \label{dialogue1}
  \centering
  \begin{tabular}{lrrrr}
    \toprule
    Model & Cuisine goal & Area goal & Price goal & Joint goal  \\
    \midrule
    Word Candidates & 82.5 ($\pm$2.2)  & 87.9 ($\pm$1.2) & 90.8 ($\pm$0.5) & 68.2 ($\pm$1.8) \\
       $+$Pretrained embeddings & 81.8 ($\pm$0.5) & 88.1 ($\pm$0.6) & 91.6 ($\pm$1.3) & 67.8 ($\pm$0.7)\\
       SLU Candidates & 77.3 ($\pm$1.1) & 89.4 ($\pm$0.5)& 91.6 ($\pm$0.3) & 65.3 ($\pm$0.9) \\
        $+$Pretrained embeddings & 78.3 ($\pm$1.7) & 89.8 ($\pm$0.8) & 91.0 ($\pm$0.8) & 65.9 ($\pm$1.2)\\
    \bottomrule
  \end{tabular}
\end{table}

\begin{table}[h]
  \caption{Accuracy result of four different training runs for ensembled
    system. Note that ~\cite{liu2018dialogue} also report ensemble results for
    their system. Our method outperforms ~\cite{rastogi2018multi} who use a similar
  candidate set ranking approach.}
  \label{dialogue2}
  \centering
  \begin{tabular}{lrrrr}
    \toprule
    Model & Cuisine goal & Area goal & Price goal & Joint goal  \\
    \midrule
    \cite{rastogi2018multi} & -  & - & - & 70.3 \\
    \cite{liu2018dialogue}  & 84 & 90  & 94 & 72 \\
    \cite{mrkvsic2016neural} & - & - & - & 73.4 \\
    \midrule
    Word Candidates & 84.0  & 88.8 & 91.7 & 70.7 \\
       $+$Pretrained embeddings & 83.0 & 89.2 & 91.8 & 70.3\\
       SLU Candidates & 78.7 & 90.3& 92.5 & 67.5 \\
        $+$Pretrained embeddings & 80.5& 91.0 & 91.6 & 68.6\\
%    Word Candidates & \textbf{84.0}  & 88.8 & 91.7 & \textbf{70.7} \\
%       $+$Pretrained embeddings & 83.0 & 89.2 & 91.8 & 70.3\\
%       SLU Candidates & 78.7 & 90.3& \textbf{92.5} & 67.5 \\
%        $+$Pretrained embeddings & 80.5& \textbf{91.0} & 91.6 & 68.6\\
    \bottomrule
  \end{tabular}
\end{table}

Our ensemble system (Table~\ref{dialogue2}) is able to outperform the work by
\citet{rastogi2018multi}. Their work, similar to us, does not have closed
candidate sets for slot values but they use sequence tagging over utterances
rather than making binary decisions. We are able to achieve competitive
performance to the hierarchical RNN system of \citet{liu2018dialogue} but our
results are not as good as the neural belief tracker
\cite{mrkvsic2016neural}. These two approaches assume closed sets for slot
values, which works for the DSTC2 data set since it does not suffer from the OOV
issue.

Note that, unlike the previous work, we do not use ASR $n$-best or in our work
as our motivation is to show the feasibility of this approach.  We observe that
the system which does joint SLU + state tracking ($n$-gram candidates) performs
better than using the SLU results as our candidate set. This could be due the
errors made by the SLU system which we are able to recover by jointly doing
state stacking and SLU~(SLU results in the data are noisy). We also observe that
using pretrained embeddings gives us a boost in performance in some of the
settings. Ensembling improves results across the board, we observe that the
experiments with higher standard deviation in Table~\ref{dialogue1} have higher
gains after ensembling.

\section{Related Work} \label{sec:related}

Dialogue state tracking (or belief tracking) aims to maintain a distribution
over possible dialogue states~\citep{bohus2006k, Williams2007partially}, which
are often represented as a set of key-value pairs. The dialogue states are then
used when interacting with the external back-end knowledge base or action
sources and in determining what the next system action should be. Previous work
on dialogue state tracking include rule-based approaches~\citep{wang2013simple},
Bayesian networks~\citep{thomson2010bayesian}, conditional random
fields (CRF)~\citep{lee2013recipe}, recurrent neural
networks~\citep{henderson2014word}, and end-to-end memory
networks~\citep{PerezLiu2017}.

In DSTC2, many systems that rely on delexicalization, where slot values from a
semantic dictionary are replaced by slot labels, outperformed systems that rely
on SLU outputs. Furthermore, the set of possible states is limited. However,
these approaches do not scale to real applications, where one can observe rich
natural language that includes previously unseen slot value mentions and large,
possibly unlimited space of dialogue states. To deal with the first issue,
\citet{mrkvsic2016neural} proposed the neural belief tracker approach that also
eliminates the need for language understanding by directly operating on the user
utterance and integrating pre-trained word embeddings to deal with the richness
in natural language.  However, their approach also does not scale to large
dialogue state space as they iterate over all slot-value pairs in their ontology
to make a decision. % provide reason More recently, \citet{rastogi2017scalable}
have proposed a candidate set ranking approach, where the candidates are
generated from language understanding system's hypotheses to deal with the
scalability issues. Our approach is the most similar to this work.  However,
their approach does not consider multi-valued slots due to the softmax layer
over all the values, whereas our approach can estimate probabilities for
multiple possible values.

Previous work that investigated joint language understanding and dialogue state
tracking include work by \citet{liu2017end, rastogi2018multi}.
\citet{liu2017end} use a hierarchical recurrent neural network to represent
utterances and dialogue flow, and estimates a probability for all possible
values, and hence suffers from the scalability issues.

\section{Conclusions} \label{sec:conclusion}

In this work we have presented a dialogue state tracking framework which is domain
agnostic. Our system is independent of the number of slots and assigns state
values using a flexible candidate set. Our system is scalable and our output is
not limited to the slot values we have seen at the training time. We show the
feasibility of our approach on the DSTC2 state tracking challenge where we
achieve competitive performance to the state-of-the-art models which do state
prediction over limited vocabulary. We think that such a domain agnostic
component will be an important step towards general-purpose end-to-end
task-oriented dialogue system. Going forward, we would like to experiment with
more data sets and system set ups. One future direction of this work is to
predict start and end of slot value token in utterances. This will relax the
n-gram restriction and allow us to handle slot values of arbitrary length
However, ultimately, we would like our system to move away with explicit state
tracking and use such a system to maintain belief over values in a latent
manner.

%% to the number of slots 

%% In this work, we have shown that it is possible to design state tracking without
%% explicit domain knowledge.

%% This is a step towards general-purpose end-to-end task-oriented dialogue system.

%% In the future, we can take a step further such that there is no explicit state
%% tracking at all --- encoding them in latent manner such that we can extract the value
%% when needed (when the system needs to call recommend restaurant API).

\bibliographystyle{acl_natbib}
\bibliography{refs}

\begin{thebibliography}{}

\end{thebibliography}


\begin{thebibliography}{26}
\expandafter\ifx\csname natexlab\endcsname\relax\def\natexlab#1{#1}\fi

\bibitem[{Bohus and Rudnicky(2006)}]{bohus2006k}
Dan Bohus and Alex Rudnicky. 2006.
\newblock A k hypotheses+ other belief updating model.
\newblock In \emph{Proc. of the AAAI Workshop on Statistical and Empirical
  Methods in Spoken Dialogue Systems}, volume~62.

\bibitem[{Bojanowski et~al.(2016)Bojanowski, Grave, Joulin, and
  Mikolov}]{bojanowski2016enriching}
Piotr Bojanowski, Edouard Grave, Armand Joulin, and Tomas Mikolov. 2016.
\newblock Enriching word vectors with subword information.
\newblock \emph{arXiv:1607.04606}.

\bibitem[{Devlin et~al.(2018)Devlin, Chang, Lee, and
  Toutanova}]{devlin2018bert}
Jacob Devlin, Ming-Wei Chang, Kenton Lee, and Kristina Toutanova. 2018.
\newblock Bert: Pre-training of deep bidirectional transformers for language
  understanding.
\newblock \emph{arXiv preprint arXiv:1810.04805}.

\bibitem[{Dhingra et~al.(2016)Dhingra, Li, Li, Gao, Chen, Ahmed, and
  Deng}]{dhingra2016towards}
Bhuwan Dhingra, Lihong Li, Xiujun Li, Jianfeng Gao, Yun-Nung Chen, Faisal
  Ahmed, and Li~Deng. 2016.
\newblock Towards end-to-end reinforcement learning of dialogue agents for
  information access.
\newblock \emph{arXiv preprint arXiv:1609.00777}.

\bibitem[{Hemphill et~al.(1990)Hemphill, Godfrey, and
  Doddington}]{hemphill1990atis}
Charles~T Hemphill, John~J Godfrey, and George~R Doddington. 1990.
\newblock The {ATIS} spoken language systems pilot corpus.
\newblock In \emph{Proceedings of the DARPA speech and natural language
  workshop}, pages 96--101.

\bibitem[{Henderson et~al.(2014{\natexlab{a}})Henderson, Thomson, and
  Williams}]{henderson2014second}
Matthew Henderson, Blaise Thomson, and Jason~D Williams. 2014{\natexlab{a}}.
\newblock The second dialog state tracking challenge.
\newblock In \emph{SIGDIAL Conference}, pages 263--272.

\bibitem[{Henderson et~al.(2013)Henderson, Thomson, and
  Young}]{henderson2013deep}
Matthew Henderson, Blaise Thomson, and Steve Young. 2013.
\newblock Deep neural network approach for the dialog state tracking challenge.
\newblock In \emph{Proceedings of the SIGDIAL 2013 Conference}, pages 467--471.

\bibitem[{Henderson et~al.(2014{\natexlab{b}})Henderson, Thomson, and
  Young}]{henderson2014word}
Matthew Henderson, Blaise Thomson, and Steve Young. 2014{\natexlab{b}}.
\newblock Word-based dialog state tracking with recurrent neural networks.
\newblock In \emph{Proceedings of the 15th Annual Meeting of the Special
  Interest Group on Discourse and Dialogue (SIGDIAL)}, pages 292--299.

\bibitem[{Hori et~al.(2018)Hori, Perez, Higasinaka, Hori, Boureau, Inaba,
  Tsunomori, Takahashi, Yoshino, and Kim}]{hori2018overview}
Chiori Hori, Julien Perez, Ryuichi Higasinaka, Takaaki Hori, Y-Lan Boureau,
  Michimasa Inaba, Yuiko Tsunomori, Tetsuro Takahashi, Koichiro Yoshino, and
  Seokhwan Kim. 2018.
\newblock Overview of the sixth dialog system technology challenge: Dstc6.
\newblock \emph{Computer Speech \& Language}.

\bibitem[{Kingma and Ba(2014)}]{kingma2014adam}
Diederik~P Kingma and Jimmy Ba. 2014.
\newblock {ADAM}: A method for stochastic optimization.
\newblock \emph{arXiv:1412.6980}.

\bibitem[{Lee and Eskenazi(2013)}]{lee2013recipe}
Sungjin Lee and Maxine Eskenazi. 2013.
\newblock Recipe for building robust spoken dialog state trackers: Dialog state
  tracking challenge system description.
\newblock In \emph{Proceedings of the SIGDIAL 2013 Conference}, pages 414--422.

\bibitem[{Liu et~al.(2017)Liu, Tur, Hakkani-Tur, Shah, and Heck}]{liu2017end}
Bing Liu, Gokhan Tur, Dilek Hakkani-Tur, Pararth Shah, and Larry Heck. 2017.
\newblock End-to-end optimization of task-oriented dialogue model with deep
  reinforcement learning.
\newblock \emph{arXiv:1711.10712}.

\bibitem[{Liu et~al.(2018)Liu, Tur, Hakkani-Tur, Shah, and
  Heck}]{liu2018dialogue}
Bing Liu, Gokhan Tur, Dilek Hakkani-Tur, Pararth Shah, and Larry Heck. 2018.
\newblock Dialogue learning with human teaching and feedback in end-to-end
  trainable task-oriented dialogue systems.
\newblock \emph{arXiv preprint arXiv:1804.06512}.

\bibitem[{Mrk{\v{s}}i{\'c} et~al.(2017)Mrk{\v{s}}i{\'c}, S{\'e}aghdha, Wen,
  Thomson, and Young}]{mrkvsic2016neural}
Nikola Mrk{\v{s}}i{\'c}, Diarmuid~O S{\'e}aghdha, Tsung-Hsien Wen, Blaise
  Thomson, and Steve Young. 2017.
\newblock Neural belief tracker: Data-driven dialogue state tracking.
\newblock In \emph{55th Annual Meeting of the Association for Computational
  Linguistics (ACL)}.

\bibitem[{Perez and Liu(2016)}]{PerezLiu2017}
Julien Perez and Fei Liu. 2016.
\newblock Dialog state tracking, a machine reading approach using memory
  network.
\newblock \emph{arXiv preprint arXiv:1606.04052}.

\bibitem[{Peters et~al.(2018)Peters, Neumann, Iyyer, Gardner, Clark, Lee, and
  Zettlemoyer}]{peters2018deep}
Matthew~E Peters, Mark Neumann, Mohit Iyyer, Matt Gardner, Christopher Clark,
  Kenton Lee, and Luke Zettlemoyer. 2018.
\newblock Deep contextualized word representations.
\newblock \emph{arXiv preprint arXiv:1802.05365}.

\bibitem[{Rastogi et~al.(2018)Rastogi, Gupta, and
  Hakkani-Tur}]{rastogi2018multi}
Abhinav Rastogi, Raghav Gupta, and Dilek Hakkani-Tur. 2018.
\newblock Multi-task learning for joint language understanding and dialogue
  state tracking.
\newblock In \emph{Proceedings of the 19th Annual SIGdial Meeting on Discourse
  and Dialogue}, pages 376--384.

\bibitem[{Serban et~al.(2016)Serban, Sordoni, Bengio, Courville, and
  Pineau}]{serban2016building}
Iulian~V Serban, Alessandro Sordoni, Yoshua Bengio, Aaron Courville, and Joelle
  Pineau. 2016.
\newblock Building end-to-end dialogue systems using generative hierarchical
  neural network models.
\newblock In \emph{AAAI}.

\bibitem[{Shao et~al.(2017)Shao, Gouws, Britz, Goldie, Strope, and
  Kurzweil}]{shao2017generating}
Louis Shao, Stephan Gouws, Denny Britz, Anna Goldie, Brian Strope, and Ray
  Kurzweil. 2017.
\newblock Generating high-quality and informative conversation responses with
  sequence-to-sequence models.
\newblock \emph{arXiv:1701.03185}.

\bibitem[{Sutskever et~al.(2014)Sutskever, Vinyals, and
  Le}]{sutskever2014sequence}
Ilya Sutskever, Oriol Vinyals, and Quoc~V Le. 2014.
\newblock Sequence to sequence learning with neural networks.
\newblock In \emph{NIPS}, pages 3104--3112.

\bibitem[{Thomson and Young(2010)}]{thomson2010bayesian}
Blaise Thomson and Steve Young. 2010.
\newblock Bayesian update of dialogue state: A {POMDP} framework for spoken
  dialogue systems.
\newblock \emph{Computer Speech \& Language}, 24(4):562--588.

\bibitem[{Wang and Lemon(2013)}]{wang2013simple}
Zhuoran Wang and Oliver Lemon. 2013.
\newblock A simple and generic belief tracking mechanism for the dialog state
  tracking challenge: On the believability of observed information.
\newblock In \emph{Proceedings of the SIGDIAL 2013 Conference}, pages 423--432.

\bibitem[{Williams et~al.(2016)Williams, Raux, and
  Henderson}]{williams2016dialog}
Jason Williams, Antoine Raux, and Matthew Henderson. 2016.
\newblock The dialog state tracking challenge series: A review.
\newblock \emph{Dialogue \& Discourse}, 7(3):4--33.

\bibitem[{Williams and Young(2007)}]{Williams2007partially}
Jason~D Williams and Steve Young. 2007.
\newblock Partially observable markov decision processes for spoken dialog
  systems.
\newblock \emph{Computer Speech \& Language}, 21(2):393--422.

\bibitem[{Yao et~al.(2016)Yao, Peng, Zweig, and Wong}]{yao2016attentional}
Kaisheng Yao, Baolin Peng, Geoffrey Zweig, and Kam-Fai Wong. 2016.
\newblock An attentional neural conversation model with improved specificity.
\newblock \emph{arXiv:1606.01292}.

\bibitem[{Zhao and Eskenazi(2016)}]{zhao2016towards}
Tiancheng Zhao and Maxine Eskenazi. 2016.
\newblock Towards end-to-end learning for dialog state tracking and management
  using deep reinforcement learning.
\newblock \emph{arXiv:1606.02560}.

\end{thebibliography}

\end{document}